\documentclass[]{interact}

\usepackage{epstopdf}
\usepackage[caption=false]{subfig}

\usepackage[longnamesfirst,sort]{natbib}
\bibpunct[, ]{(}{)}{;}{a}{,}{,}
\renewcommand\bibfont{\fontsize{10}{12}\selectfont}

\theoremstyle{plain}

\theoremstyle{definition}

\theoremstyle{remark}

\begin{document}

\articletype{ARTICLE TEMPLATE}

\title{A Perceptron-based Fine Approximation Technique for Linear Separation}

\author{
\name{Ákos Hajnal\textsuperscript{a,b}\thanks{CONTACT Ákos Hajnal. Email: akos.hajnal@sztaki.hu} }
\affil{\textsuperscript{a}Institute for Computer Science and Control (SZTAKI), Hungarian Research Network (HUN-REN), Budapest, Hungary; \textsuperscript{b}Óbuda University, John von Neumann Faculty of Informatics, Budapest, Hungary}
}

\maketitle

\begin{abstract}
This paper presents a novel online learning method that aims at finding a separator hyperplane between data points labelled as either positive or negative. Since weights and biases of artificial neurons can directly be related to hyperplanes in high-dimensional spaces, the technique is applicable to train perceptron-based binary classifiers in machine learning. In case of large or imbalanced data sets, use of analytical or gradient-based solutions can become prohibitive and impractical, where heuristics and approximation techniques are still applicable. The proposed method is based on the Perceptron algorithm, however, it tunes neuron weights in just the necessary extent during searching the separator hyperplane. Due to an appropriate transformation of the initial data set we need not to consider data labels, neither the bias term. respectively, reducing separability to a one-class classification problem. The presented method has proven converge; empirical results show that it can be more efficient than the Perceptron algorithm, especially, when the size of the data set exceeds data dimensionality. 
\end{abstract}


\section{Introduction}

Artificial neural networks are constructed of several tens of thousands of neurons (or even more) these days, organized into layers or blocks, having various activation functions and connections (feed-forward, recurrent, residual, etc.) aiming at realizing some complex classification or regression task \citep{sadeeq2020}. Such large neural networks are typically trained using a variant of the Stochastic Gradient Descent (SGD) method with Back-propagation \citep{robbins1951, haskell1944, rumelhart1986}, combined with some regularization and hyperparameter tuning. This approach has not merely proved to be applicable to a wide range of architectures including convolutional neural networks but achieved unprecedented results in various domains such as image recognition, speech synthesis, language translation, self-driving cars, etc. 

SGD is a general optimizer that can efficiently tune neural network parameters (neuron weights and biases), however, the final result, which is a trained model, is often extremely difficult or impossible to interpret. Assessment on a test set (a separated portion of input data unseen during training) of the obtained "black box" model provides moderate information on the robustness or reliability. It raises concerns at applying such neural networks to critical real-life tasks (e.g., in health care), and so there is an increasing demand for meaningful explanation, interpretation of the constructed machine learning models.
Geometrical interpretation is one way to approach this problem. Support Vector Machines (SVM) \citep{vapnik1995,boser1992,cortes1995} are noted representatives that aim at finding the optimal separating hyperplane with large margin.

Data separation is an intrinsic behaviour of artificial neurons in neural networks; neuron weights and biases can directly be related to hyperplanes in high-dimensional spaces. Use of various activation functions might make difficult imagine or visualize this geometrical representation, but even having a single hidden layer with any non-linear neuron activation function (e.g., signum, which provides only the information on which side of the hyperplane the data point resides), the neural network is still a universal approximator \citep{cybenko1989,hornik1989, csaji2001}. Also, classifiers implemented as a deep neural network contain neurons in the output layer each corresponding to a category, which aim at firing 1 if the input belongs to a particular class or 0 otherwise (one-hot encoding). Those neurons perform pure separation tasks on data points obtained from the previous layers.

This paper presents an online learning method that aims at finding a separator hyperplane between positive and negative data samples, and so applicable to train perceptron-based binary classifiers in supervised machine learning. Heuristics and approximation techniques such as the one proposed are still applicable in case of large or imbalanced data sets when analytical or gradient-based solutions might become infeasible or impractical due to resource requirements. 

The main contributions of the paper are:
\begin{itemize}
\item[--] a three-step data transformation technique that eliminates the need of considering data labels and bias term during search,
\item[--] an iterative heuristic to determine the separator hyperplane, which updates its orientation in just the necessary extent,
\item[--] empirical results on large data sets and its comparison with the Perceptron algorithm.
\end{itemize}

The rest of the paper is structured as follows. In section 2, we overview existing techniques and approaches. In section 3, we present a data transformation technique, then, in section 4, a method to determine (or approximate) the separator hyperplane. In section 5, our preliminary experimental results are shown. Finally, section 6 concludes the paper.

\section{Related work}

The first computational models for artificial neurons were introduced by McCulloch and Pitts \citep{mcculloch1943} and Rosenblatt  \citep{rosenblatt1957}. An artificial neuron, given a set of weight values \textbf{w} of size \textit{d} and a bias \textit{b}, performs the computation 
\( \sum_{i=1}^{d} {w_i}{x_i} + {b} \) 
on input vector \textbf{x} having \textit{d} features 
(\( \mathbf{w}, \mathbf{x} \in \mathbb{R}{^d} \)), 
which is eventually transformed by a non-linear activation function to get the output value of the neuron.
\( \sum_{i=1}^{d} {w_i}{x_i} \) equals to the dot product: \textbf{wx}. \textbf{wx} + \textit{b} = 0 gives the equation of a hyperplane \textit{H} in the \textit{d}-dimensional space, where input vector \textbf{x} represents a point. \textbf{wx} + \textit{b} returns a value proportional to the signed Euclidean distance of \textbf{x} from \textit{H}. Assuming sign activation function (\(sgn\)), the neuron simply determines on which side of \textit{H} \textbf{x} resides, i.e., returns +1 if \textbf{x} is on the positive side (the half space to which normal vector \textbf{w} points to) or –1 otherwise.

In binary classification, we are given a set of \textit{n} data samples: \(X = (\textbf{x}_1, \textbf{x}_2, \ldots, \textbf{x}_n)\) where each data sample \(\textbf{x}_i\) is labelled by value \(y_i\), which is +1 (positive sample) or 
–1 (negative sample), indicating the class that \(\textbf{x}_i\) belongs to. Training a neuron is the process of determining values \textbf{w} (weights) and \textit{b} (bias) such that \(sgn(\textbf{w}\textbf{x}_i + b)\) correctly predicts \(y_i\) for (ideally) all data samples. Geometrically, it is equivalent to finding a separator hyperplane with normal vector \textbf{w} and bias \textit{b} that passes in-between the positive and negative data points in a \textit{d}-dimensional space, having all positive samples on one side and all negatives on the other, respectively.

The separation problem above can be addressed by analytical methods such as linear programming (refer to \citep{walk2022} for an overview), assigning an inequality for each data sample, and solving them for \textbf{w} and \textit{b}. When the size of the data set is large, this method however requires considerable computational resources, system memory. In such cases, online training methods \citep{hoi2021} are often preferred that gradually improves the result considering one sample at a time. 

 Another approach is to use Support Vector Machines (SVM) \citep{fan2008,chang2011,chandra2021}. These solutions aim at computing the optimal, a maximal-margin hyperplane with linear kernel, equal distance from both the positive and negative samples. To obtain large margin is often not required, when we are only interested in separation, and SVMs performance might degrade when the amounts of positive and negative samples are imbalanced \citep{nunez2017}.
 
Feed-forward neural networks (Multi-layer Perceptrons, MLPs) are also often applied to solve data separation or classification (e.g., \citep{keup2022,mohammadazadeh2022,lin2022,jodoin2023}). These are more complex architectures composed of several neurons (trained using SGD) that typically cannot be associated with a single linear decision boundary (i.e., linear separability). Using linear activation function these networks collapse to a single neuron, regardless of the number of neurons or layers. 

Rosenblatt presented the Perceptron algorithm in 1957, which is an online training method for binary classifiers, using a single neuron. It follows a mistake-driven approach that updates current weight vector \(\textbf{w}^t\) (initially zero) with data vector \(\mathbf{x}_i\), if its label is wrongly predicted by current \(\textbf{w}^t\) (i.e., \(sgn(\textbf{w}^t\textbf{x}_i) \neq y_i \)), obtaining an updated weight vector \(\textbf{w}^{t+1}\):

\begin{equation}
\textbf{w}^{t+1} = \textbf{w}^t + y \cdot \textbf{x}_i
\end{equation}

Factor \textit{y} is +1 on false negatives or $-$1 on false positives, respectively. This algorithm gradually improves the weight vector, tolerates imbalanced data, and is proven to reach convergence (solution) in finite steps. The number of required updates is bounded by \(R^2/\sigma^2\), where \textit{R} is the length of maximal data vector (maximum of \( \lVert \mathbf{x}_i \rVert \)), and \( \sigma \) is the size of a “margin” between the positive and negative data points (Euclidean distance of the nearest points from the optimal separating hyperplane). Value \(\sigma\) is typically not known in advance, which can be a very small value, therefore, even after a large number of updates, non-separability cannot be stated with certainty. If given a data set with only two data: 100 and 101, where 100 is a negative sample and 101 is a positive sample (we have a single feature), the Perceptron algorithm requires more than 20,000 epochs to complete. It is because the algorithm undercorrects or overcompensates current \(\textbf{w}^t\) most of the times with the (full) value of update vector \(\textbf{x}_i\). The method presented in this work will require only a single update to solve this problem.

Numerous variants of the Perceptron algorithm have been evolved in the past decades that aim at improving the its robustness, finding large margin separator or being applicable to non-separable datasets, respectively. Pocket algorithm \citep{gallant1990}, Maxover algorithm \citep{wendemuth1995}, Winnow algorithm \citep{littlestone1988}, the Voted Perceptron \citep{freund1998}, to name a few.

\section{Data pre-processing}

In this section, we present three data transformation steps to be done prior searching for the separating hyperplane. These operations are computationally cheap, require no global information about the dataset (e.g., maximum/minimum values), and so can potentially be performed on-the-fly (considering a single data sample at a time).

As a first step (\textit{dimension extension}) we extend each data sample with an additional (constant) feature, which will allow us not to deal with bias term at searching the hyperplane (only hyperplanes passing the origin need to be investigated). In the second step (\textit{negative inversion}), we negate all feature values in each negative sample in order to avoid the need to consider data labels of the data points. As a final step (\textit{data normalization}), we normalize data samples to have length 1 (\( \forall \lVert \mathbf{x}_i \rVert = 1 \))), which only aims at simplifying calculations and the proof of convergence.

\subsection{Dimension extension}

To find a separating hyperplane defined by equation \textbf{wx} + \textit{b} = 0 we need determine its normal vector \textbf{w} (orientation) and bias \textit{b} (offset). It is a common trick to avoid searching for the bias \textit{b} to add an additional feature to each data sample corresponding to constant value: 1, and also extend \textbf{w} with one more element. 

If a data sample \textbf{x} has \textit{d} features initially: \(\mathbf{x} = (x_1, x_2, \ldots, x_d)\), then the expanded feature vector \(\mathbf{x}'\) will be: \(\mathbf{x'} = (x_1, x_2, \ldots, x_d, 1)\). If we also extend normal vector \textbf{w}, and denote its last element as \textit{b}, we obtain obtain: \(\mathbf{w}' = (w_1, w_2, \ldots, w_d, b)\). The dot product \(\mathbf{w'x'}\) yields \textbf{wx} + \textit{b} · 1, which corresponds to the value computed by the original hyperplane. Therefore, if we find an orientation \(\mathbf{w}'\) in the \textit{d}+1 dimensional space of hyperplane: \(\mathit{H'}\): \(\mathbf{w'x'}\) = 0 (without bias) that separates all transformed data \(\mathbf{x}'\), we also get a separating hyperplane \textit{H}: \textbf{wx} + \textit{b} = 0 in the original data space, where bias \textit{b} is the last element of vector \(\mathbf{w}'\) and \textbf{w} can be obtained from \(\mathbf{w}'\) by simply removing its last element.

Note that this transformation does not reduce the number of parameters to find, but merely makes search easier to consider only the orientation (rotation). 

\subsection{Negative inversion}

The dimension extension trick described above lets us search for a "bias-less" hyperplanes: \textbf{wx} = 0 that yields positive value for positive samples and negative value otherwise. Note that in this section we denote by \textbf{x} the data sample already dimension extended and \textbf{w} has one more parameters that the number of features the original data set. 

To avoid the need to consider data labels, we transform further the data samples in a way that we take the opposite of all feature values in all negative samples. That is, if \(\mathbf{x} = (x_1, x_2, \ldots, x_d, 1)\) is a negative sample, then it is replaced by a feature vector \(\mathbf{x'} = (-x_1, -x_2, \ldots, -x_d, -1)\) in the data set -- leaving feature values of positive samples unchanged. 

After this second transformation, our task further reduces to finding a hyperplane that has all data points on one side regardless of the original labels (one-class classification). If we have a hyperplane \textbf{wx} = 0 that yields positive value \(\textbf{wx}>0\) for all the transformed data points (i.e., all on one side), and \textbf{x} was originally labelled as a negative sample, then \(\mathbf{wx'} = \mathbf{w}({-1} \cdot \mathbf{x}) = {-}\mathbf{wx}\) will result in a negative value,  whereas \(\mathbf{wx'}\) 
yields positive values for all (unchanged) positive samples. Therefore, \textbf{wx} = 0 is indeed a separator hyperplane of the original data set, returning positive value for positive samples and negative value for negative samples. 

\begin{figure}
\centering
\resizebox*{7.5cm}{!}{\includegraphics{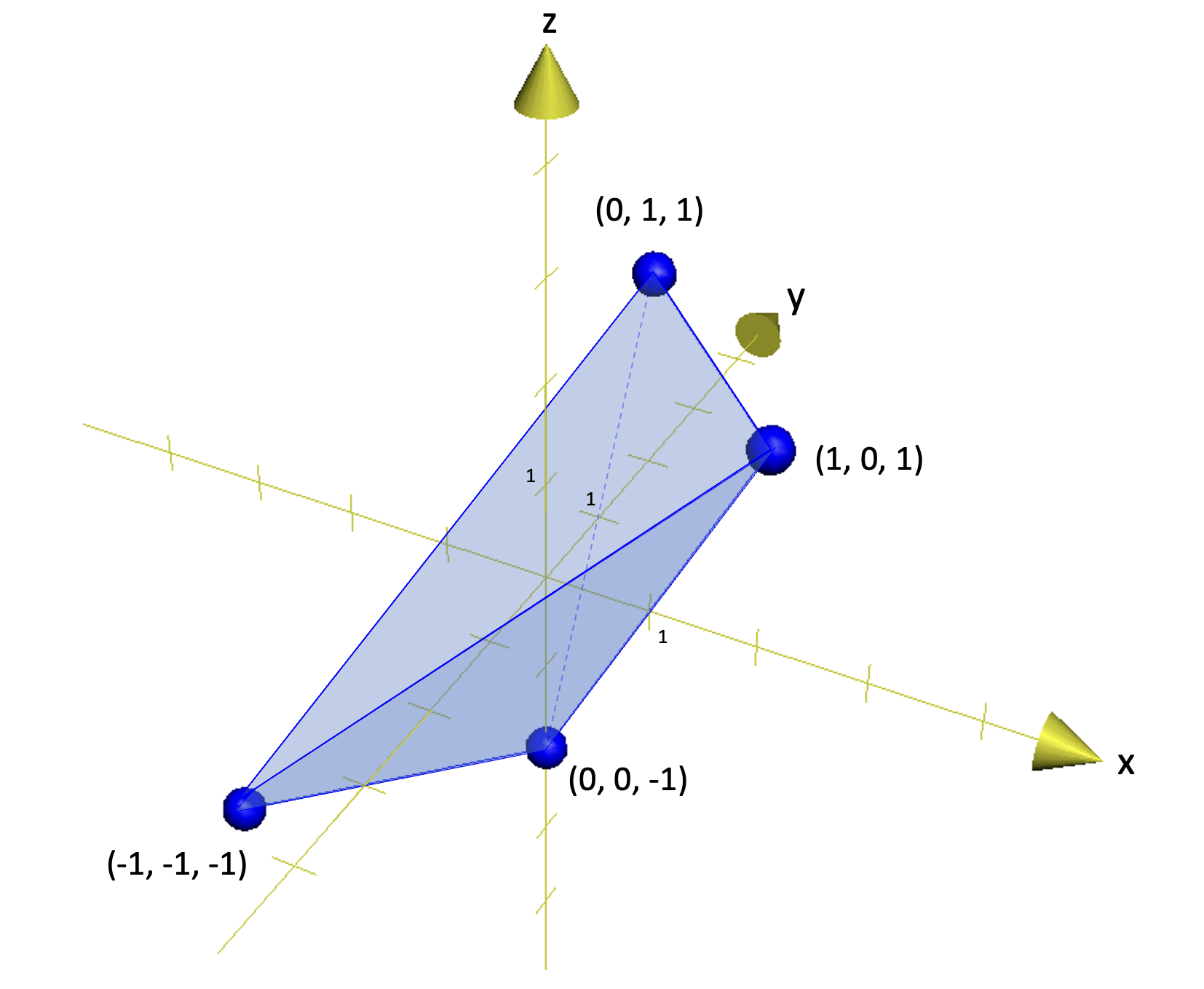}}
\hspace{5pt}
\caption{Convex hull of the XOR problem.} 
\label{figure1}
\end{figure}

As an illustration, let us consider the XOR problem and follow the data transformation steps. The original data set is composed of two positive samples: (0, 1), (1, 0) and two negatives: (0, 0), (1, 1). After dimension extension, we obtain positives: (0, 1, 1), (1, 0, 1) and negatives: (0, 0, 1), (1, 1, 1). Inversion results in (unlabelled) data points: (0, 1, 1), (1, 0, 1), (0, 0, -1), (-1, -1, -1). These point are illustrated in Figure~\ref{figure1}. As it might be seen, there exists no separator hyperplane that has all the points on one side (the convex hull of these points contains the origin), therefore the original data set is not separable.

\subsection{Data normalization}

The last transformation step is simply to normalize all data samples in the training set \(X\) (already dimension extended and negative inverted) by dividing each feature value of $\mathbf{x}_i$ with the Euclidean length of the encompassing data vector $\lVert\mathbf{x}_i\rVert$. As a result, all data samples in the transformed data set \(X'\) will have length 1, i.e., \(\lVert \mathbf{x'}_i \rVert = 1, \forall \mathbf{x}_i \in X \)). 

Intuitively, this transformation performs a projection of data points to a unit-radius hypersphere in the \textit{d+1}-dimensional space (considering dimension extension). Since it means a division of data vectors with a positive constant \textit{c} (note that all dimension extended data samples have length $\geq 1$), it has no effect on separability (sign of \(\textbf{w}(c \cdot \textbf{x})\) corresponds to \(\textbf{wx}\)). 

The problem we are given now is a unit sphere populated with a set of points, and our goal is to slice it with a single hyperplane passing through the origin such that only one of its halves contains data points.
Note that the original data set is separable if and only if the transformed data set is separable, and if we find a separating hyperplane, the hyperplane separating the original data set can easily be derived.

\section{Perceptron approximation}

The approximation method presented in this section is similar to the Perceptron algorithm: it is an online learning approach that processes one data sample at a time and mistake-driven that updates on failed predictions only. It assumes data set transformed in advance as described in the previous section, starts from an initial hyperplane with normal vector \textbf{w}, and will gradually improve its orientation whenever we find another data that is falsely predicted (\textbf{wx} evaluates to a negative number, i.e., lies on the wrong side of the hyperplane). Recall that our goal is to find a hyperplane passing through the origin having all data points on the positive side.  

The difference from the Perceptron algorithm is that how we update \textbf{w}. The Perceptron algorithm adds \textbf{x} to or subtracts \textbf{x} with its full value from current \textbf{w}, respectively, which though rotates \textbf{w} to the "proper direction": \textbf{wx} decreases on false positives and increases on false negatives, but often undercorrects (still having wrong \textbf{wx} value, with incorrect sign) or overcompensate (having correct \textbf{wx} but greater than necessary). Our update rule corrects current \(\mathbf{w}^{t}\) in just the necessary extent to make \(\mathbf{w}^{t+1}\mathbf{x} = 0\), which can be formalized as follows: 

\begin{equation}
\label{eq0}
\mathbf{w}^{t+1} = \mathbf{w}^{t} - (\mathbf{w}^{t}\mathbf{x})\mathbf{x}
\end{equation} 

\begin{figure}
\centering
\resizebox*{7.5cm}{!}{\includegraphics{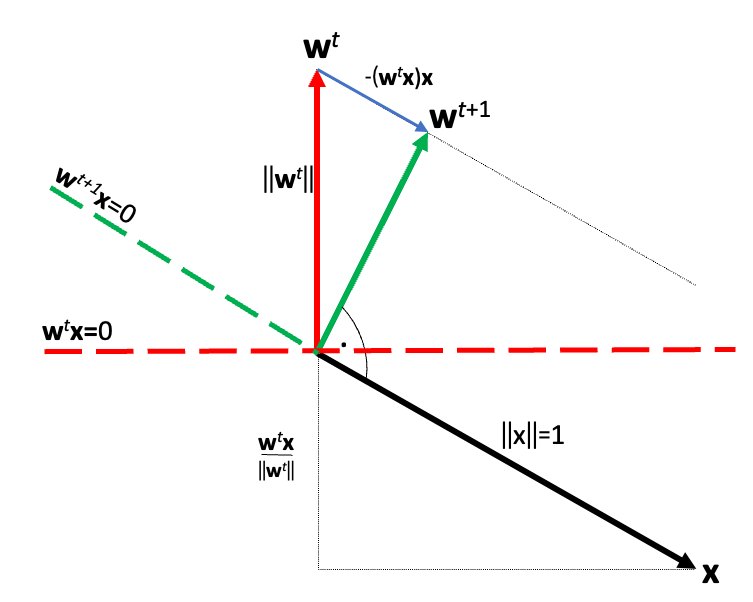}}
\hspace{5pt}
\caption{Geometrical interpretation of the update rule.} 
\label{figure2}
\end{figure}

Intuitively, it means a rotation of the normal vector to the direction of data point \textbf{x} till it becomes perpendicular to \textbf{x}, and so \textbf{x} will lie exactly on the new hyperplane defined by \(\mathbf{w}^{t+1}\), as illustrated in Figure~\ref{figure2}.

To see why the addition of vector \textbf{x} multiplied by factor \(-\mathbf{w}^{t}\mathbf{x}\) gives zero dot product for $\mathbf{w}^{t+1}$ and data input \textbf{x} we can write: 

\begin{equation}
 \begin{split}
\mathbf{w}^{t+1}\mathbf{x} = (\mathbf{w}^{t} - (\mathbf{w}^{t}\mathbf{x}) \cdot \mathbf{x})  \mathbf{x} =  
\mathbf{w}^{t}\mathbf{x} - (\mathbf{w}^{t}\mathbf{x}) \cdot \mathbf{x} \mathbf{x} = 
\\
= \mathbf{w}^{t}\mathbf{x} - (\mathbf{w}^{t}\mathbf{x}) \cdot \lVert \mathbf{x} \rVert ^{2} =
\mathbf{w}^{t}\mathbf{x} - (\mathbf{w}^{t}\mathbf{x}) \cdot 1 = 0
\end{split}
\end{equation}

In the following paragraphs, we show that the method gets closer to a solution at each update. Assume that the data set is separable and there is a separator hyperplane \( \mathbf{w}^{*} \mathbf{x} = 0 \) with normal vector \(\mathbf{w}^{*}\) that has all the data samples on its positive side (\(\mathbf{w^{*}x} > 0 \) for all input data \textbf{x} in \(X\)). Also assume a unit-long normal vector for \(\mathbf{w}^{*}\), i.e., \(\lVert\mathbf{w}^{*}\rVert = 1\) (without loss of generality). Let us investigate how the dot product of current normal vector \(\mathbf{w}^{t}\) and "ideal" normal vector \(\mathbf{w}^{*}\) changes over the updates:

\begin{equation}
\label{eq1}
\mathbf{w}^{t+1}\mathbf{w^{*}} = (\mathbf{w}^{t} - \mathbf{w}^{t}\mathbf{x} \cdot \mathbf{x} )\mathbf{w^{*}} =  
\mathbf{w}^{t}\mathbf{w}^{*} - \mathbf{w}^{t}\mathbf{x} \cdot \mathbf{w^{*}x} 
\end{equation}

Since \(\mathbf{w}^{t}\mathbf{x}\) evaluates to a negative number (\(\mathbf{w}^{t}\) needs update) and \(\mathbf{w}^{*}\mathbf{x}\) is positive, the value \(-\mathbf{w}^{t}\mathbf{x}\cdot\mathbf{w}^{*}\mathbf{x}\)  will be positive. Therefore:

\begin{equation}
\label{eq1b}
\mathbf{w}^{t+1}\mathbf{w}^{*} > \mathbf{w}^{t}\mathbf{w}^{*} 
\end{equation}

The square of the length of the new normal vector \(\mathbf{w}^{t+1}\) (after update) can be calculated as follows:
\[
\lVert \mathbf{w}^{t+1} \rVert ^2 = (\mathbf{w}^{t} - \mathbf{w}^{t}\mathbf{x}\cdot \mathbf{x})^2 =
\lVert \mathbf{w}^{t} \rVert ^2 + (\mathbf{w}^{t}\mathbf{x})^2 \cdot \lVert \mathbf{x} \rVert ^2 - 2 \cdot \mathbf{w}^{t} \cdot (\mathbf{w}^{t}\mathbf{x}) \cdot \mathbf{x} 
\]
\begin{equation}
\label{eq2}
=\lVert \mathbf{w}^{t} \rVert ^2 + (\mathbf{w}^{t}\mathbf{x})^2 \cdot 1 - 2 \cdot \mathbf{w}^{t}\mathbf{x}^2 =
\lVert \mathbf{w}^{t} \rVert ^2 - (\mathbf{w}^{t}\mathbf{x})^2
\end{equation}

Consequently, the square of the length of the new normal vector \(\lVert \mathbf{w}^{t+1} \rVert ^2\) will be less than \(\lVert \mathbf{w}^{t} \rVert ^2\)  by the square of dot product \( \mathbf{w}^{t}\mathbf{x} \), therefore:

\begin{equation}
\label{eq2b}
\lVert \mathbf{w}^{t+1} \rVert < \lVert \mathbf{w}^{t} \rVert
\end{equation}
Dot product \( \mathbf{w^{t+1}w^{*}} = \lVert \mathbf{w}^{t+1} \rVert  \cdot \lVert \mathbf{w}^{*} \rVert \cdot cos(\alpha^{t+1})\), where \(\alpha^{t+1}\) is the angle between normal vector \( \mathbf{w}^{t+1} \) and the "ideal" normal vector \( \mathbf{w^{*}} \). \(\lVert \mathbf{w}^{*} \rVert = 1\), therefore \( \mathbf{w^{t+1}w^{*}} = \lVert \mathbf{w}^{t+1} \rVert \cdot cos(\alpha^{t+1})\). 
Similarly, \( \mathbf{w^{t}w^{*}} = \lVert \mathbf{w}^{t} \rVert \cdot cos(\alpha^{t})\),  where \(\alpha^{t}\) is the angle in the \textit{t}th step.

From \( \mathbf{w}^{t+1}\mathbf{w}^{*} > \mathbf{w}^{t}\mathbf{w}^{*} \) 
(equation (\ref{eq1b})) it follows that: 

\begin{equation}
\lVert \mathbf{w}^{t+1} \rVert \cdot cos(\alpha^{t+1}) > \lVert \mathbf{w}^{t} \rVert \cdot cos(\alpha^{t})
\end{equation}

Since 
\(\lVert \mathbf{w}^{t+1} \rVert < \lVert \mathbf{w}^{t} \rVert\) 
(equation (\ref{eq2b})), \( cos(\alpha^{t+1}) \) must be greater 
than \( cos(\alpha^{t}) \). Cosine is decreasing in (0, $\pi$) and \(
0^{\circ} < \alpha^{t}, \alpha^{t+1}< 180^{\circ}\), therefore: 

\begin{equation}
\alpha^{t+1} < \alpha^{t}
\end{equation}

This means that if we apply update rule (\ref{eq0}), in each iteration, the current normal vector gets closer to the normal vector of an actual separator hyperplane, i.e., the method converges.

Note that despite strict monotonic convergence the approximation might slow down when false positives or false negatives get very close to the current hyperplane, and so they produce little values for $\mathbf{wx}$, and little updates on $\textbf{w}$. Therefore, the method does not guarantee finite step termination.

When implementing this solution, even at using double precision floating-point numbers, $\mathbf{w}^{t+1}\mathbf{x}$ will often not exactly be zero, but a very small value -- occasionally positive or even negative, respectively, in the latter case, still producing false prediction. Therefore, it is preferable to use a small positive constant $\epsilon$ in the update rule to avoid such cases: 

\begin{equation}
\label{eq3}
\mathbf{w}^{t+1} = \mathbf{w}^{t} + (\epsilon - \mathbf{w}^{t}\mathbf{x}) \cdot \mathbf{x}
\end{equation}

As a result, $\mathbf{w}^{t+1}\mathbf{x}$ will yield correctly a non-negative value after an each update. 

\begin{figure}
\centering
\resizebox*{8cm}{!}{\includegraphics{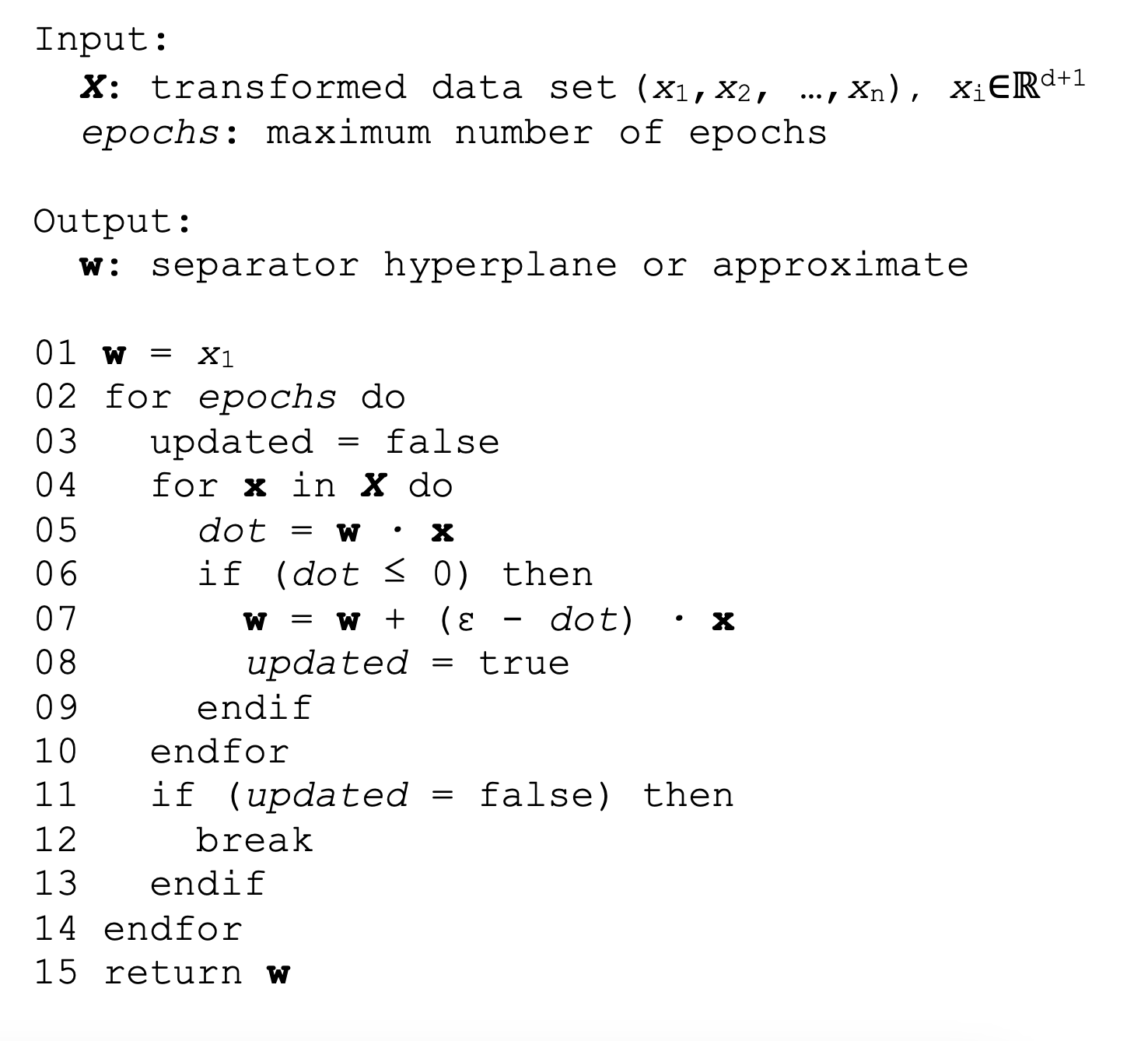}}
\hspace{5pt}
\caption{Pseudo-code of the algorithm.} 
\label{figure3}
\end{figure}

The pseudocode the approximation algorithm is shown in Figure~\ref{figure3}. The procedure expects a data set, which has already been dimension extended, negative inverted and normalized (as described in the previous section). In line 1, we initialize the normal vector $\mathbf{w}$ to point to the first element of the data set (could also be chosen at random). Line 2 repeats iterating over all data elements for a specified number of epochs. We reset updated flag (\textit{updated}) before each epoch in line 3. We iterate over each data sample in the data set, and for each we compute the dot product of the current normal vector and the data sample (line 5). If it proves to be non-positive (line 6), we apply the update rule on $\mathbf{w}$ (line 7); also set updated flag (line 8). After each epoch, we check whether update has happened (line 11), and if not, we are done and can exit from the loop (line 12). If no update has happened during the last epoch, the procedure returns the normal vector of a separating hyperplane having all transformed data points on one side, i.e. the solution. In case we reach the number of allowed epochs, we get an approximate solution. 

\section{Empirical Results}

It easy to see that the proposed method successfully solves the "100--101" separation problem, achieving the result, a separating hyperplane with a single update. To construct a more challenging task, we artificially generated data sets with "small" margin and with offset. First, a hyperplane is created with a randomly selected normal vector (each element is picked from range (-1, 1) using uniform distribution) and with zero bias, then we generated the specified number of data samples (using the same distribution) and labeled them depending on which the side of the hyperplane they reside. Finally, we shifted all data points along a randomly selected bias vector. In this way, we could get a separable, biased and labelled data set in any given dimensions and with any data set size. 

We compared the Perceptron algorithm and the presented approximation heuristic. Both methods get the same data set in each run for the sake of fair comparison. In the implementation of the Perceptron algorithm, we used no data normalization. We were interested in how they behave for relatively large data sets in high dimensions, which are also challenging for analytical methods (e.g., linear programming).

\begin{figure}
\centering
\subfloat[Accuracy over epochs of the (P)erceptron algorithm and the (A)pproximation method (10,000 samples, 100,000 dimensions, 3 runs).]{%
\resizebox*{7cm}{!}{\includegraphics{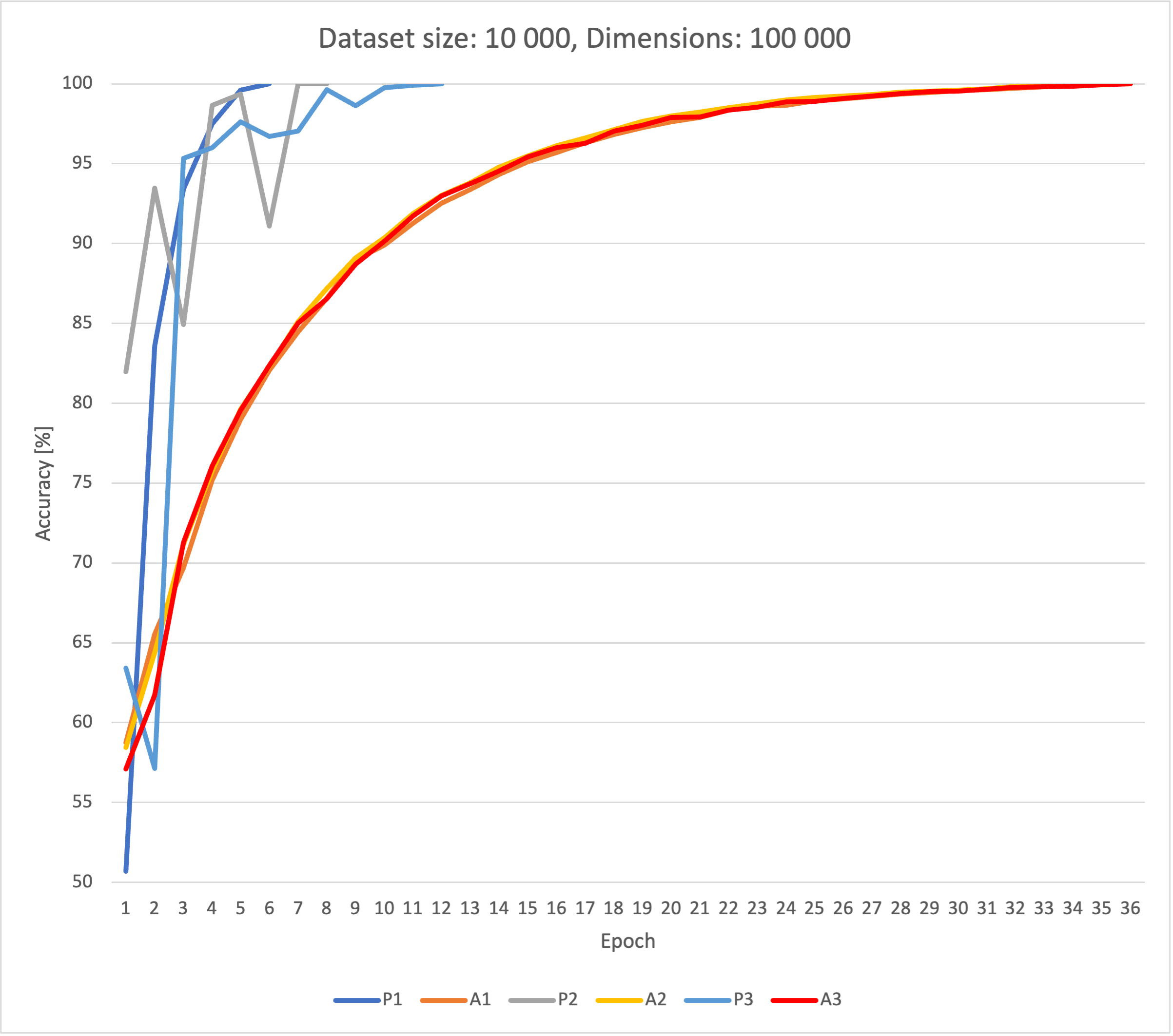}}}\hspace{5pt}
\subfloat[Accuracy over epochs of the (P)erceptron algorithm and the (A)pproximation method (100,000 samples, 10,000 dimensions, 3 runs).]{%
\resizebox*{7cm}{!}{\includegraphics{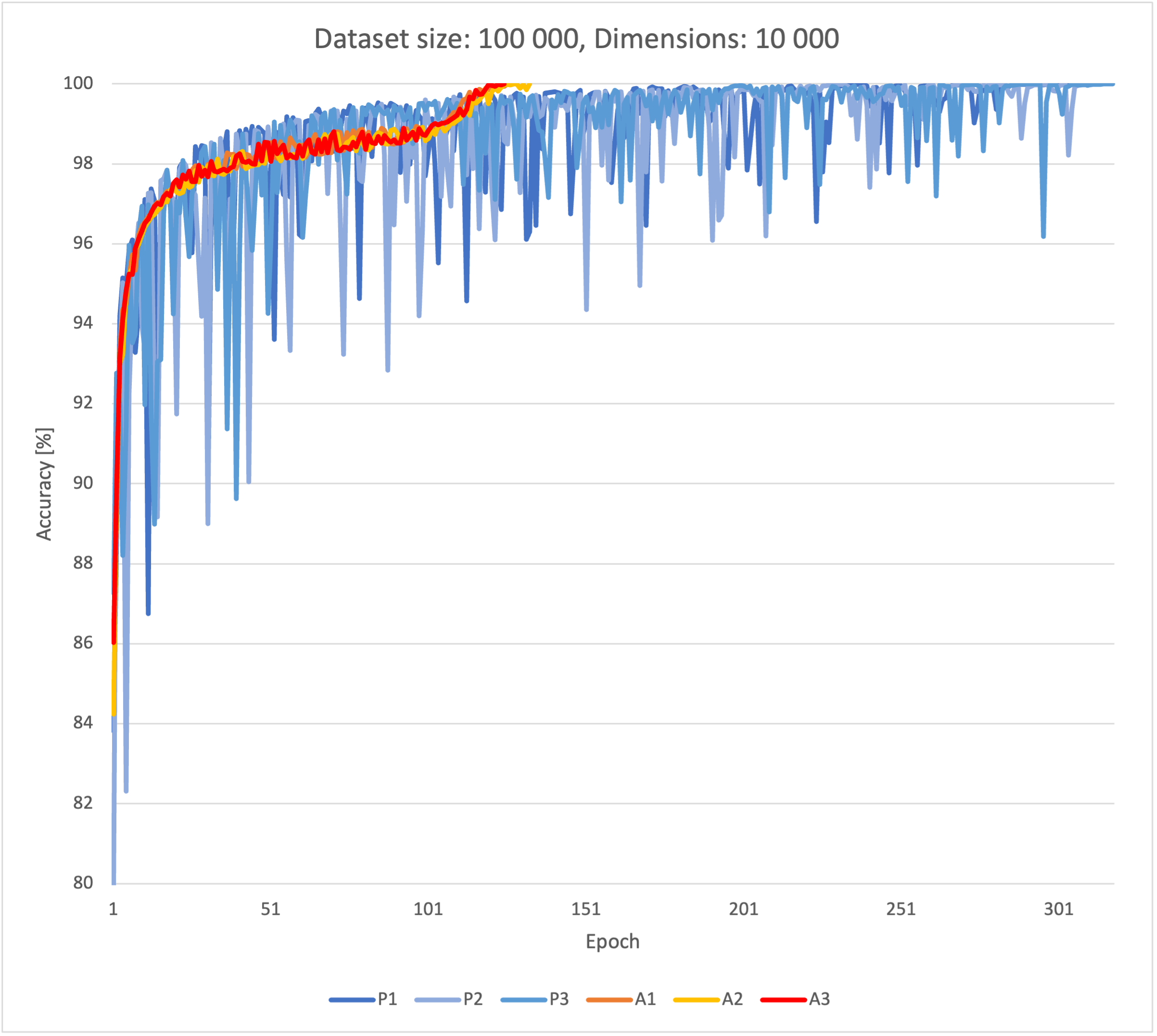}}}

\subfloat[Accuracy over epochs of the (P)erceptron algorithm and the (A)pproximation method (1,000,000 samples, 100 dimensions).]{%
\resizebox*{7cm}{!}{\includegraphics{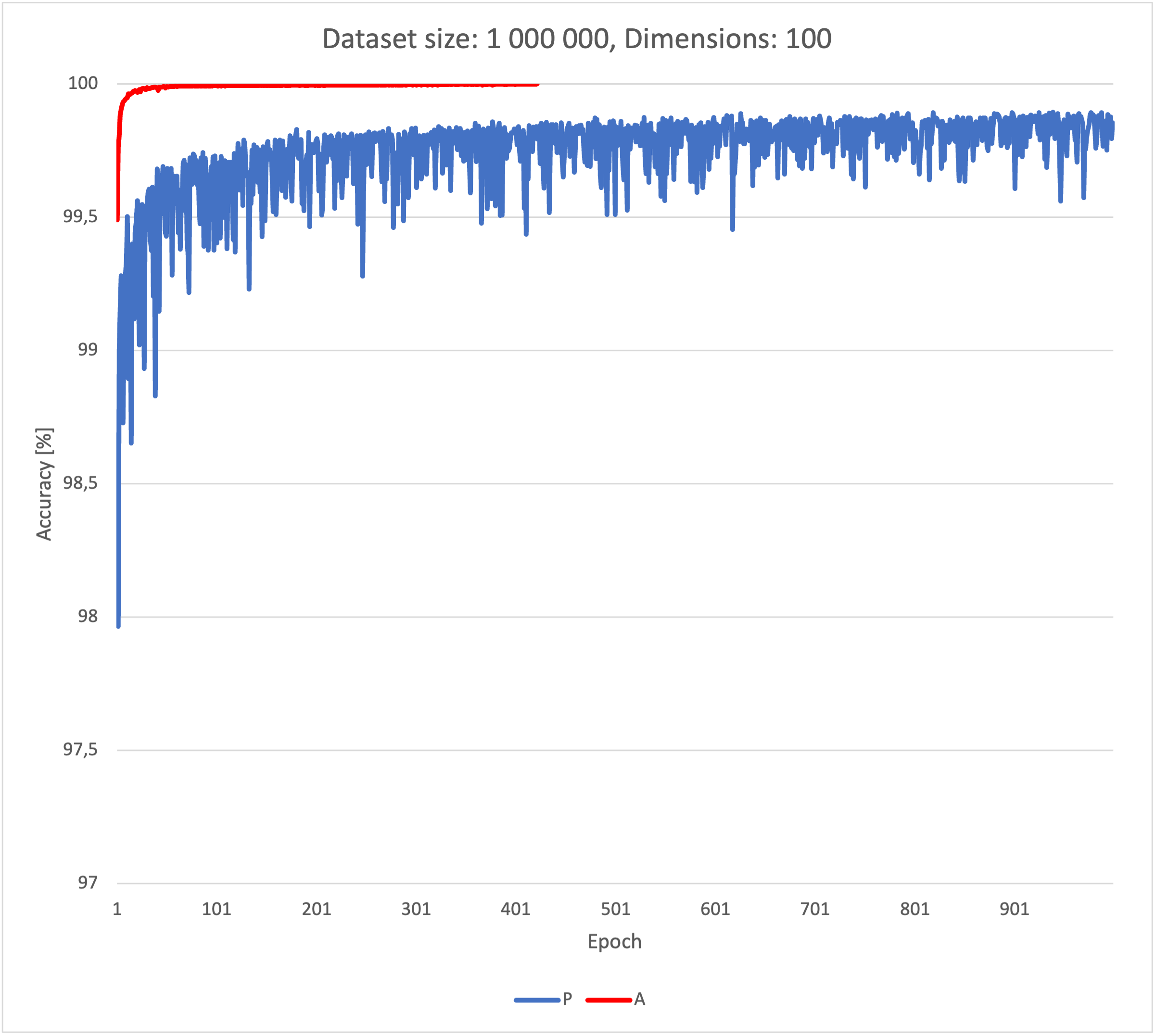}}}\hspace{5pt}
\caption{Comparision of the performance of the Perceptron algorithm and the Approximation heuristic.} 
\label{charts}
\end{figure}

In the first experiment, we chose a relatively small data set size: 10,000 and high dimensions: 100,000, and we measured how the accuracy (number of correct predictions per all samples) evolved during the epochs (limited to 1000). We expected that the "Approximation" method will show fast convergence due to its precise update technique. The results are shown in Figure \ref{charts}.a, where “P” denotes the Perceptron algorithm, “A” denotes the Approximation method (the number following this letter shows the experiment number; three runs have been done; each on newly generated data sets).

The results did not confirm our expectations: the Perceptron algorithm was faster (completed in less epochs: 12 vs. 36) and converged more quickly to accuracy 100\%. The only slight advantage of the approximation technique was its steady and smooth increase. 

The performance of the methods however changed when data set size was much greater than the number of dimensions, data features. In \ref{charts}.b, the results of the next experiment are shown, which was carried out on 100,000 data samples and in 10,000 dimensions. As shown in the chart, we can see that the approximation technique was twice faster as the Perceptron this time -- still without significant fluctuation. 

The difference is even more significant when we have 1 million test data with only 100 features, for which the Perceptron couldn’t find a solution within 1000 epochs, while the approximation technique converged quickly and terminated with the solution within 500 epochs. It is shown in \ref{charts}.c. 

\section{Conclusions}

This work presented a novel, lightweight, online learning method that aimed at finding a separator hyperplane between positive and negative data samples. Hyperplanes and neuron weights are directly related, and so the presented method can be applied to train perceptron-based binary classifiers in supervised learning.  

The technique iteratively adjusts the orientation of an initial randomly selected hyperplane, in each step in just the necessary extent. It has been proven that such updates converge to the solution -- though without a guarantee of finite step termination. Due to data transformations performed prior to search it needs not to determine consider the bias neither to consider data labels, respectively, during search, still capable of providing a solution for the original (non-transformed) data set at the end.

Experimental results showed that the method could efficiently be applied when the number of data samples exceeded the number of features (data dimensionality), so it could be a lightweight alternative of other online learning techniques.

The method quickly approximates the solution but slows down when a set of data points are near to the decision boundary, trying to fit to one of the faces of the hull from outside. It needs further investigation how this “fine tuning” phase can be sped up or avoided. Also, it is an interesting research question how the technique can be improved and “regularized” to find a hyperplane with greater margin from data points.

\end{document}